\documentclass[a4paper, 10pt, conference]{ieeeconf}  
\IEEEoverridecommandlockouts   
\usepackage[caption=false,font=footnotesize]{subfig}
\usepackage[T1]{fontenc}
\usepackage[latin9]{inputenc}
\usepackage{color}
\usepackage{xcolor,colortbl}
\usepackage{textcomp}
\usepackage{graphicx}
\usepackage{array,booktabs}
\usepackage{algorithm,algorithmicx,algpseudocode}
\usepackage{amsmath} 
\usepackage{amssymb}  
\usepackage{multirow}
\usepackage{epstopdf}
\usepackage{siunitx}
\usepackage{mathtools}
\usepackage{multicol}
\usepackage{tabularx}
\usepackage{hhline}
\usepackage{xfrac}
\usepackage[11pt]{moresize}
\pdfmapfile{=fullembed.map}

\title{\LARGE \bf
	3D Surface Reconstruction from Voxel-based Lidar Data 
}

\author{Luis Rold\~ao$^{1,2}$, Raoul de Charette$^{1}$ and Anne Verroust-Blondet$^{1}$ 
	\thanks{$^{1}$Robotics and Intelligent Transportation Systems (RITS) Team, INRIA Paris, 2 Rue Simone Iff, 75012 France. {\tt\small \{raoul.de-charette, anne.verroust\}@inria.fr}.} \thanks{$^{2}$R\&D Department of AKKA Technologies. 78280 Guyancourt, France. {\tt\small \{luis.roldao@akka.eu\}}.}
}

\begin{document}
	
\maketitle
\thispagestyle{empty}
\pagestyle{empty}

\begin{abstract}

To achieve fully autonomous navigation, vehicles need to compute an accurate model of their direct surrounding. In this paper, a 3D surface reconstruction algorithm from heterogeneous density 3D data is presented.  The proposed method is based on a TSDF voxel-based representation, where an adaptive neighborhood kernel sourced on a Gaussian confidence evaluation is introduced. This enables to keep a good trade-off between the density of the reconstructed mesh and its accuracy. Experimental evaluations carried on both synthetic (CARLA) and real (KITTI) 3D data show a good performance compared to a state of the art method used for surface reconstruction.

\end{abstract}

\section{Introduction}\label{sec:introduction}

A robust and accurate model of the environment is crucial for autonomous vehicles. In fact, imprecise representations of the vehicle's surrounding may lead to unexpected situations that could endanger the passengers. Although many different geometrical representations have been proposed by both the robotics \cite{Stoyanov2013} and the graphics \cite{Berger2017} communities, creating an accurate 3D model of the surroundings from mounted sensors remains a challenge.

In recent years, the constant evolution of Lidar sensors enables to obtain rich and accurate information, even in large and complex scenes. However, the size of the input data, along with the noise, misaligned scans and density variation, make the task a very challenging one. To overcome this, coarse 3D representations are often used and restrictive hypotheses are applied (i.e. planar assumptions for road detection). While these representations can be sufficient to perform path planning or obstacle avoidance, complex maneuvers might require a more accurate description of the scene's geometry. This can be specially convenient for terrain mapping or off road navigation.

In this paper, we present an algorithm capable to perform a fine and accurate 3D surface reconstruction of the environment from depth sensors. From the statistics of the input point cloud sampled into a voxel grid, local approximations of the surface are performed using an adaptive neighborhood capable to cope with the heterogeneous density of the input scan. Then, a truncated signed distance field (TSDF) is estimated to obtain a continuous mesh that maintains a high level of detail and density in areas close to the vehicle.
Our output mesh can be of special interest for both the robotics and the graphics
community to perform different tasks, such as terrain traversability assessment or physical modeling.


\begin{figure}
	\centering
	\includegraphics[width=0.46\textwidth]{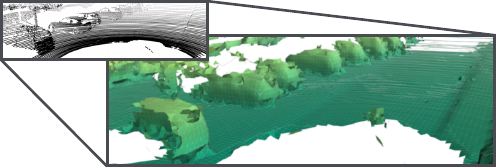}
	\caption{Our pipeline reconstructs 3D surfaces from input Lidar point clouds (here KITTI dataset), keeping a good trade-off between accuracy and mesh density. We first rely on the approximation with explicit local surfaces, and then estimate a signed distance field from which the mesh is extracted at zero-crossing.}
	\vspace{-2.5mm}
	\label{fig:intro_figure}
\end{figure}

\begin{figure*}
	\centering
	\includegraphics[width=0.99\textwidth]{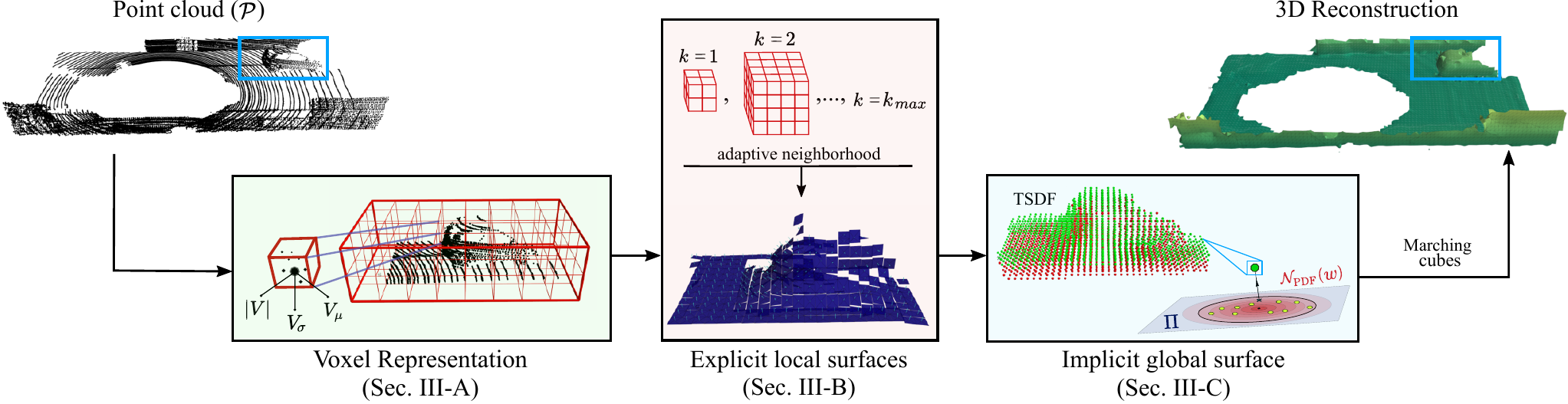}
	\caption{Overview of our method. From left to right: the point cloud $\mathcal{P}$; the voxel grid at where the statistical distribution of the points is updated (Sec. \ref{subsec:voxel_representation}); local surface approximations at neighborhoods $k \in [1, k_{max}]$ of the grid vertices (Sec. \ref{subsec:neighborhood_definition}); TSDF calculated from the planes by considering its confidences from the statistics distribution (Sec. \ref{subsec:implicit_global_surface}); final 3D reconstruction.}
	\label{fig:pipeline}
\end{figure*}

\section{Related work}\label{sec:related_work}

The 3D geometry of a scene described from range sensors can be represented in many different ways. The choice of the representation is often related to its final purpose: visualization, mapping, localization, terrain traversability, among many others. 

Some methods use directly the 3D set of points received from the input sensor, which might be useful for visualization or obstacle detection tasks. However, the level of details of the representation depends on the amount of data used, that rapidly becomes prohibitive for outdoor scenes. Conversely, other works propose a regularly sampled grid as introduced in \cite{Moravec1985}, where occupancy information is stored into each cell. This enables to handle big amounts of data more efficiently and reduce memory needs, specially by using recursive structures such as octrees \cite{Bares1989, Hornung2013}. These approaches have become widely used for terrain traversability assessment, mapping and visualization. However, their discrete nature does not enable a continuous representation, which might be desirable for other tasks such as physical modeling.

Alternatively, the graphics community has explored different methods to create a triangular mesh from the 3D points of the scanned surface. These methods have a wide range of applications on completely different fields as described in \cite{Berger2017}. For simplicity, we distinguish between explicit and implicit methods.

\textit{Explicit methods:} These techniques are either parametric or triangulated with respect to how they represent the reconstructed surface. For some applications, the continuity of the reconstruction is not required. Some researchers propose to fit local primitive shapes to the input point cloud and obtain a segmented mesh with simple primitives for the surface description \cite{Vanegas2012, Schnabel2007}. These approaches are not suitable for reconstructing complex scenes. 

Other methods create local descriptions of the surface by a set of unoriented discs (-\textit{surfels}-) calculated from the points distribution inside defined neighborhoods. Surfels have been used for traversability assessment \cite{Ryde2013}, and more recently to perform simultaneous localization and mapping (SLAM) in outdoor urban environments \cite{Behley2018}.

\textit{Implicit methods:} For other applications such as physical modeling or detailed terrain traversability, a more accurate and continuous representation might be preferred. To obtain this, some approaches commonly define a Truncated Signed Distance Field (TSDF) to represent the surface implicitly by a gradient field \cite{Berger2017}. This implicit representation needs to be post-processed by meshification algorithms (e.g. Marching Cubes \cite{Lorensen1987}) to obtain the final mesh.

In \cite{CurlessLevoy1996}, range information across different viewpoints are integrated to average a TSDF from where the scalar field is obtained. By using this technique, 3D modeling has been performed in both small indoor \cite{Newcombe2011} and large scale outdoor scenes \cite{Whelan2012, Steinbrucker2013}. These methods typically require a large number of viewpoints to output a dense reconstruction and are susceptible to outliers.

Other implicit methods perform local approximations of the surface from where the TSDF is obtained. Poisson reconstruction \cite{Kazhdan2006, Kazhdan2013} is a well-known technique for creating watertight surfaces from oriented point samples. Point-based methods such as Hoppe's \cite{Hoppe1992} or Implicit Moving Least Squares (IMLS) \cite{Shen2004, Kolluri2008, bouchiba2019} locally fit the data to a lower degree polynomial by using a projection operator. For such methods, variable density inputs might represent a challenge since the data is treated equally along the complete space.

In contrast to most methods presented above, that are unable to accommodate to the heterogeneous density of the input data while keeping the accuracy of the reconstruction, we introduce a method that is able to handle this variable density while keeping a good trade-off between the accuracy and the density of the outputted surface.

\section{Methodology}\label{sec:methodology}

Let us consider an input point cloud $\mathcal{P}$ obtained from any range sensor at a known viewpoint and sampled by a voxel grid, our aim is to reconstruct the underlying 3D mesh.
From the statistical distributions of the points in each voxel (Sec.~\ref{subsec:voxel_representation}), we first approximate local planar surfaces (Sec.~\ref{subsec:explicit_local_surfaces}), and then compute the implicit surface representation (Sec.~\ref{subsec:implicit_global_surface}) that encodes the distances to the local planar surfaces.
To accommodate to the input data heterogeneous density, as well as to gain robustness to noise, we use an adaptive neighborhood kernel, resulting in a denser and smoother reconstruction.
The overall overview of our methodology is shown in Fig. \ref{fig:pipeline}.

\subsection{Voxel representation}\label{subsec:voxel_representation}

We benefit of the work of \cite{roldao2018statistical} to update efficiently a regular voxel-wise representation of the point cloud $\mathcal{P}~\rightarrow~\{V^1, V^2, ..., V^n\}$, with voxel size $\alpha$.
In addition to the number of points $|V|$, each voxel $V$ stores the 3D statistical distribution of the points lying inside, that is: the mean $V_{\mu}$ and the covariance $V_{\sigma}$.
This enables a rich compact representation of the points inside each voxel, while being significantly lighter than storing all the points.
The statistical distribution is computed incrementally upon insertion of new points.

\subsection{Explicit local surfaces}\label{subsec:explicit_local_surfaces}

It has been shown in \cite{Ryde2013, Behley2018}, that complex environments are efficiently approximated as a set of primitive local surfaces. 
This is especially suitable for mobile robotics, as robots usually evolve in well structured environments. 
Following this observation, we compute local planar surfaces using the 3D statistical distribution described in the previous section. 
While a naive implementation would fit planar surfaces to each voxel independently, this would inherently lead to a noisy reconstruction since some voxels may have very few points, if not none, due to the heterogeneous density of Lidar data.

In our pipeline we propose using an adaptive neighborhood definition where local surfaces are estimated from multi-scale voxels statistics. 
Our neighborhood definition presents two main advantages: a)~it increases the statistical robustness which improves large planar surface estimation (e.g. ground, walls), b)~it counterbalances the lack of local data due to low density or occlusion.

\subsubsection{Neighborhood definition}\label{subsec:neighborhood_definition}

We define the multi-scale neighborhood at the vertices location rather than voxels, since implicit surfaces will later be estimated at each vertex of the voxel representation. 
Let's consider $v$ a vertex from the voxel-grid representation, its 8 adjacent voxels make up the first neighborhood level denoted $\mathcal{H}^1(v)$. 
Subsequently, the union of $\mathcal{H}^1(v)$ and the voxels adjacent to $\mathcal{H}^1(v)$ make up the neighborhood $\mathcal{H}^{2}(v)$.
More formally, the neighborhood $\mathcal{H}^k(v)$ is composed of the $(2k)^3$ nearest voxels surrounding~$v$.
A two-dimensional illustration of $\mathcal{H}^k(v)$ at levels $k=1$ and $k=2$ is presented in Fig. \ref{fig:neighborhood_plane_confidence}.

Following \cite{Ryde2013}, for a given neighborhood $\mathcal{H}$ (indices are dropped for clarity) we obtain the cardinal $|\mathcal{H}|$, statistical mean $\mathcal{H}_{\mu}$, and covariance $\mathcal{H}_{\sigma}$ from the merging of statistical data of all voxels in $\mathcal{H}$.
\\

\subsubsection{Planar estimation}\label{subsec:planar_estimation}

Having obtained the local statistics, we use the covariance $\mathcal{H}_{\sigma}$ to estimate the local planar surface of $\mathcal{H}$ through a Principal Component Analysis (PCA) if the following equation is satisfied:

\begin{equation}\label{eq:minimum_number_points}
	|\mathcal{H}|~\geq~N_{min}\,,
\end{equation}
\\
where $N_{min}$ is a hyper parameter. Suppose ($\overrightarrow{e_1}, \overrightarrow{e_2}, \overrightarrow{e_3}$) the eigen vectors, and ($\lambda_1, \lambda_2, \lambda_3$) the eigen values with $\lambda_1~\geq~\lambda_2\geq~\lambda_3$, we define $\overrightarrow{e_3}$ (the least dominant axis) as the unoriented normal of the planar estimation of the surface at neighborhood $\mathcal{H}$.
Since normals need to be consistently oriented, the normal $\overrightarrow{n}$ of the plane is oriented towards the sensor pose $s_p$ as follows:

\begin{equation}\label{eq:normal_orientation}
	\overrightarrow{n} =
	\begin{dcases}
		\overrightarrow{e_3}
		& 
		\text{if $\overrightarrow{e_3}\cdot(s_p - \mathcal{H}_{\mu}) > 0$}\,,
		\\
		-\overrightarrow{e_3}
		&
		\text{otherwise}\,.
	\end{dcases}
\end{equation}

\begin{figure}
	\centering
	\includegraphics[width=0.9\columnwidth]{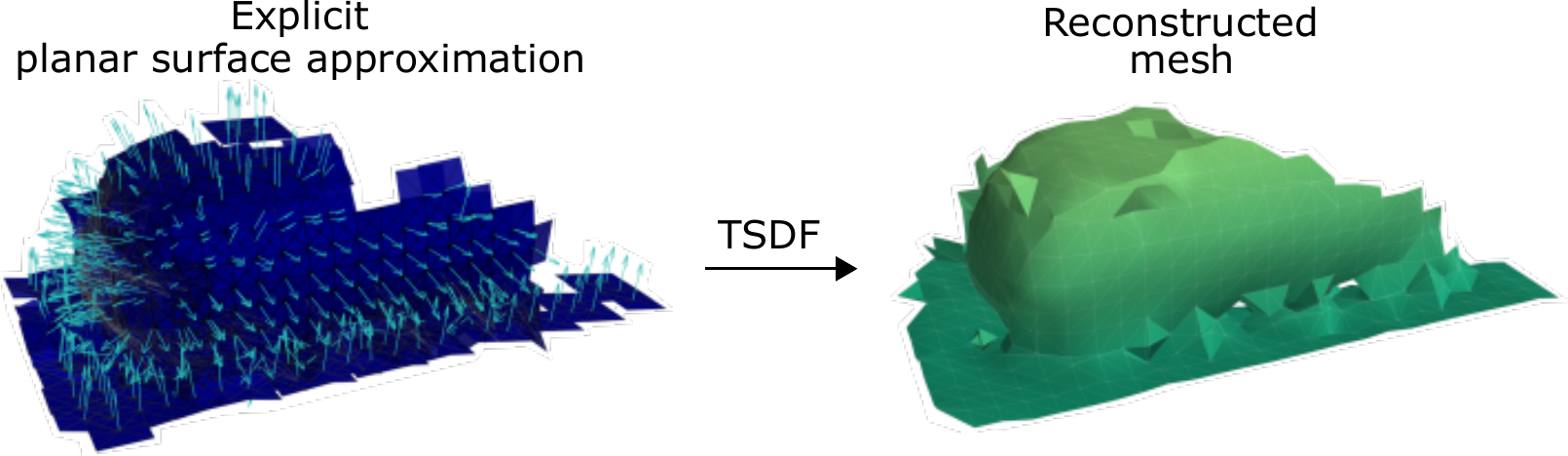}
	\caption{Explicit local planar surfaces with their estimated normals for $k=1$ (left) and its corresponding mesh reconstruction from the TSDF (right). Note that applying $k=1$ neighorhood leads to a noisy surface estimation (e.g. car edges) that is overcomed with our adaptive neighborhood strategy.}
	\label{fig:planes_different_k}
\end{figure}

\noindent{}where $\cdot$ stands for the dot product. We denote $\Pi$ the plane formed by the pair of normal and center $(\overrightarrow{n}, \mathcal{H}_{\mu})$. An example of the local planar surfaces estimation is visible in Fig.~\ref{fig:planes_different_k}.

\subsection{Implicit global surface}\label{subsec:implicit_global_surface} 

To reconstruct the global continuous surface, we compute the Truncated Signed Distance Field (TSDF) for each vertex $v \subseteq V$ close to the scanned surface. 

To cope with the varying density of points in the point cloud, we first compute an optimal neighborhood level $k'$ at each vertex, and then estimate the TSDF value given this optimal neighborhood.

\begin{figure}
	\centering
	\includegraphics[width=0.8\columnwidth]{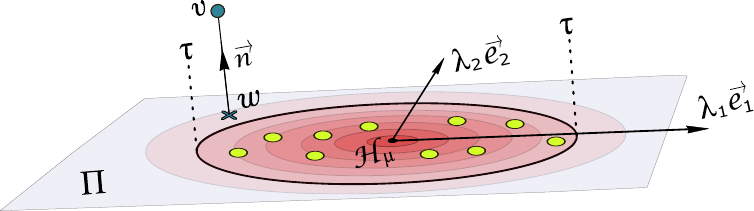}
	\caption{The likelihood of $w$ belonging to $\Pi$ is estimated from $\mathcal{N}_{\text{PDF}} (w \;|\;  \mathcal{H}_{\mu}, \Sigma)$ shown in red. The likelihood must be higher than $\tau$ in order to consider $\Pi$ as a valid plane for $w$. ($k$ indices dropped for clarity.)}
	\label{fig:neighborhood_gaussian}
\end{figure} 	

\begin{figure}
	\centering
	\includegraphics[width=0.8\columnwidth]{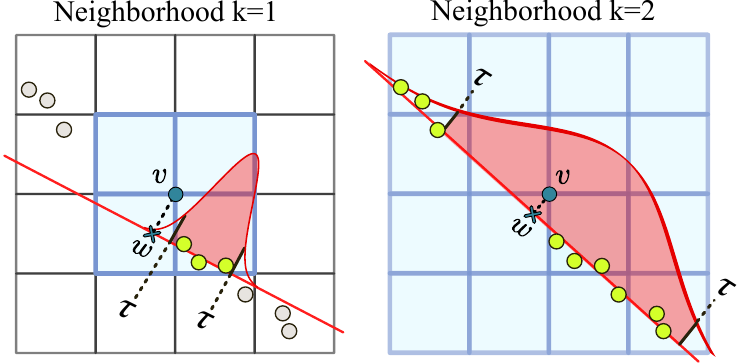}
	\caption{Analogous 2D representation of the dynamic neighborhood. Planar surface approximations are performed at different neighborhood levels $k \in [1, k_{max}]$. The considered plane corresponds to the minimum neighborhood level that satisfies Eq. \ref{eq:pdf_value}, where the schematic 1D Gaussian distribution is shown in red.}
	\label{fig:neighborhood_plane_confidence}
\end{figure}

\subsubsection{Adaptive neighborhood}\label{subsec:adaptive_neighborhood}

A naive implementation of our neighborhood definition would use a constant $k$ throughout the scene. 
However, large $k$ values will over smooth high density regions, while small values of $k$ will lead to noisy estimation in low density regions.
Instead, we compute the optimal neighborhood level $k'$ for each vertex $v$, given the probability of $v$ to belong to a multivariate Gaussian distribution projected on $\Pi^{k'}$ (the neighborhood planar estimation).

For each vertex, we calculate the projection $w^k$ of $v$ onto the plane $\Pi^k(v)$ and evaluate the likelihood of this projection to belong to the Gaussian $\mathcal{N}^k$, where $\mathcal{N}^k$ is the 2D planar-Gaussian of the statistical distribution of $\mathcal{H}^k$ projected onto $\Pi^k$, as illustrated in Fig.~\ref{fig:neighborhood_gaussian}.
The optimal neighborhood level $k'$ is defined as the smallest level for which the projection of $v$ onto $\Pi^k$ has a probability to belong to the Gaussian $\mathcal{N}^k$ greater than $\tau$ (a hyper parameter). 
Formally, it is the smallest integer for which the probability density function $\mathcal{N}_{\text{PDF}}$ satisfies 

\begin{equation}
	\label{eq:pdf_value}
	\mathcal{N}^k_{\text{PDF}} (w^k \;|\;  \mathcal{H}^k_{\mu}, \Sigma)  \geq \tau\,;  \,\, \Sigma = {\scriptsize \left[ {\begin{array}{cc} \lambda_1 & 0 \\ 0 & \lambda_2 \end{array} } \right]}\,.
\end{equation}

\noindent{}In practice, we compute the optimal $k'$ iteratively starting at level 1 and stops when the above equation is satisfied as it is shown in Fig. \ref{fig:neighborhood_plane_confidence}.
To avoid exponential computation time, $k$ is bounded between $[1, k_{max}]$.

\subsubsection{TSDF computation}\label{subsec:tsdf_computation}

We now compute the TSDF value for each vertex of the voxel grid, from the optimal local neighborhood level, while accounting for the normal estimation at the corresponding level. 
In other words, we compute $\text{TSDF}(v)$ such that:

\begin{equation}\label{eq:TSDF}
	\text{TSDF}(v) = \overrightarrow{n^{k'}} \cdot (v - \mathcal{H}^{k'}_{\mu})\,.
\end{equation}

\noindent{}Our adaptive neighborhood level selection can efficiently handle varying local density and fill gaps between missing data, such as those between two adjacent Lidar layers.
Subsequently, the condition on the probability density function given in Eq.~\ref{eq:pdf_value} avoids the extension of the surface at its borders which is visible in qualitative results.

After TSDF computation, we use the popular marching cubes \cite{Lorensen1987} to extract the mesh from the zero-crossing level of the gradient field. Fig. \ref{fig:planes_different_k} shows the reconstructed mesh with constant neighborhood ($k=1$).

\section{Experimental results}\label{sec:experimental_results}

We evaluate our proposal on synthetic data from CARLA~\cite{Dosovitskiy17} and real data from KITTI~\cite{Geiger2013}, and compare our performance against the IMLS baseline~\cite{Shen2004}.
For both synthetic and real data we used 100 frames equi-sampled from either an urban-like sequence for CARLA, or the residential sequence 0018 for the public KITTI dataset.

\subsection{Methodology}\label{subsec:methodology}

Unless stated otherwise, our hyper-parameters remain unchanged in all experiments, with: $\alpha=0.2m$, $\tau=0.2$, $N_{min}=10$, and $k_{\text{max}}=5$. 
For fair comparison, the spherical neighborhood radius of IMLS is set to $\alpha \times k_{\text{max}} = 1m$, and its k-nearest neighbor search to $N_{min}$. The noise parameter of IMLS is set to $h= \sfrac{1}{3} \, (\alpha \times k_{\text{max}}) = 0.33$.

Noteworthy, it is impractical to have a real mesh ground-truth and KITTI can only be used for qualitative evaluation.
Conversely, we use synthetic data from CARLA to evaluate the benefit of each of our contributions. 
The setup in CARLA replicates a top-mounted Velodyne HDL64E, similarly to the real KITTI setup.
We also simulate a collocated noise-free Lidar with abnormally high resolution (316~layers), which serves as ground-truth for the reconstruction.
Consequently, we frame the evaluation as a set-to-set distance problem, and measure the quality as the distance of the predicted mesh vertices (P) to the ground-truth (GT) set.

We use two metrics derived from the literature: the Average Error (AE), and the Haussdorf Distance (HD). 
The former measures the average distance error from one point to its nearest point in the other set, that is: $\displaystyle{}AE_{P\rightarrow{}GT} = \sum_{a\in{}P}\frac{1}{|P|}\min_{b\in{}GT}|a-b|$.
The Haussdorf distance~\cite{Aspert2002} is classically used for point set distances and gives a sense of the largest minimum error, that is: $\displaystyle{}\text{HD}_{P\rightarrow{}GT} = \max_{a\in{}P}\min_{b\in{}GT}|a-b|$.
As each of the two metrics are directed, we also report the symmetrical metric, as the average of both directed metrics. For Haussdorf, $\text{HD}_{\text{sym}} = 0.5(\text{HD}_{P\rightarrow{}GT}+\text{HD}_{GT\rightarrow{}P})$.
We chose not to use the Chamfer distance used in machine learning~\cite{fan2017point} because it isn't a metric-scale and is thus harder to interpret intuitively.

\begin{table}
	\caption{Performance on synthetic data}
	\centering
	\footnotesize
	\scriptsize
	\setlength{\tabcolsep}{0.017\linewidth}
	\definecolor{Gray}{gray}{0.85}
	\newcolumntype{s}{>{\columncolor{Gray}}c}
	\begin{tabular}{ll|ccs|ccs|} 
		\multicolumn{2}{c|}{} & \multicolumn{3}{c|}{\textbf{Average Error (m)}} & \multicolumn{3}{c|}{\textbf{Hausdorff Distance (m)}}  \\
		\multicolumn{2}{c|}{\textbf{Method}} & {{\scriptsize P{\tiny$\rightarrow$}GT}} & {{\scriptsize GT{\tiny$\rightarrow$}P}} & {{\scriptsize Sym}} & {{\scriptsize P{\tiny$\rightarrow$}GT}} & {{\scriptsize GT{\tiny$\rightarrow$}P}} & {{\scriptsize Sym}} \\
		\hline
		IMLS \cite{Shen2004} & & $0.37$ & ${0.09}$ & $0.23$ & $4.54$ & ${8.32}$  & $\textbf{6.43}$  \\
		\hline
		\multirow{3}{*}[0.1cm]{\textbf{{\scriptsize Ours AN+GC}}} 	& {{\tiny$\mathbf{k_{\text{max}}=5}$}} & $0.14$ & $0.13$ & $\textbf{0.14}$ & $1.39$ & $20.49$ & $10.94$ \\
		& {{\tiny$k_{\text{max}}=3$}}  & ${0.09}$ & $0.26$ & $0.17$ & $0.69$ & $30.84$ & $15.77$ \\
		\hline
		\multirow{3}{*}[0.1cm]{{\scriptsize Ours AN}} 		& {{\tiny$k_{\text{max}}=5$}}   & $0.30$ & $0.12$ & $0.21$ & $1.69$ & $20.36$ & $11.02$ \\
		& {{\tiny$k_{\text{max}}=3$}}   & $0.14$ & $0.25$ & $0.19$ & $0.87$ & $30.83$ & $15.85$ \\
		\hline
		\multirow{3}{*}[0.0cm]{{\scriptsize Ours CN+GC}} 	& {{\tiny$k=5$}}         & $0.15$ & $0.16$ & $0.15$ & $1.38$ & $20.50$ & $10.94$ \\
		& {{\tiny$k=3$}}         & ${0.09}$ & $0.27$ & $0.18$ & $0.69$ & $30.85$ & $15.77$ \\
		& {{\tiny$k=1$}}        & ${0.03}$ & $3.44$ & $1.73$ & ${0.24}$ & $65.28$ & $32.76$\\
		\hline
		\multirow{3}{*}[0.0cm]{{\scriptsize Ours CN}} 	& {{\tiny$k=5$}}         & $0.30$ & $0.14$ & $0.22$ & $1.69$ & $20.36$ & $11.03$ \\
		& {{\tiny$k=3$}}         & ${0.14}$ & $0.26$ & $0.20$ & $0.87$ & $30.83$ & $15.85$ \\
		& {{\tiny$k=1$}}        & ${0.03}$ & $3.44$ & $1.73$ & ${0.24}$ & $65.28$ & $32.76$\\
		\hline		
	\end{tabular}
	\label{tab:ablation_results}
\end{table}

\begin{figure}
	\centering
	\scriptsize
	\setlength{\tabcolsep}{0.017\linewidth}
	\renewcommand{\arraystretch}{0.8}
	\begin{tabular}{cc}
		\textbf{Ours AN+GC} & Ours AN\\
		\includegraphics[width=0.47\columnwidth]{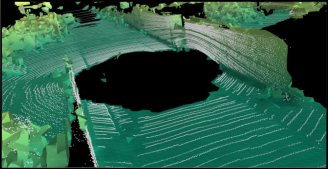} & \includegraphics[width=0.47\columnwidth]{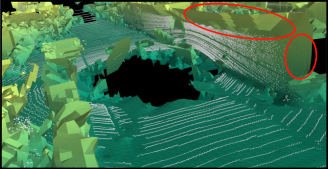}\\
		Ours CN+GC & Ours CN\\
		\includegraphics[width=0.47\columnwidth]{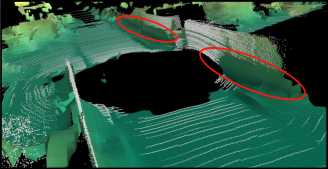} & \includegraphics[width=0.47\columnwidth]{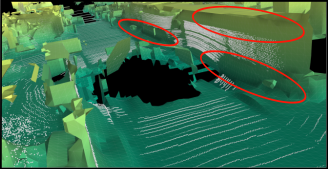}\\
	\end{tabular}
	
	\caption{Qualitative comparison of our method by removing the main components of our proposal. Images correspond to a single frame of a point cloud from CARLA. In shown images $k$ or $k_{max}$ equals $5$.}
	\label{fig:ablation_comparison}
\end{figure}

\begin{figure*}[!h]
	\centering
	\includegraphics[width=0.83\textwidth]{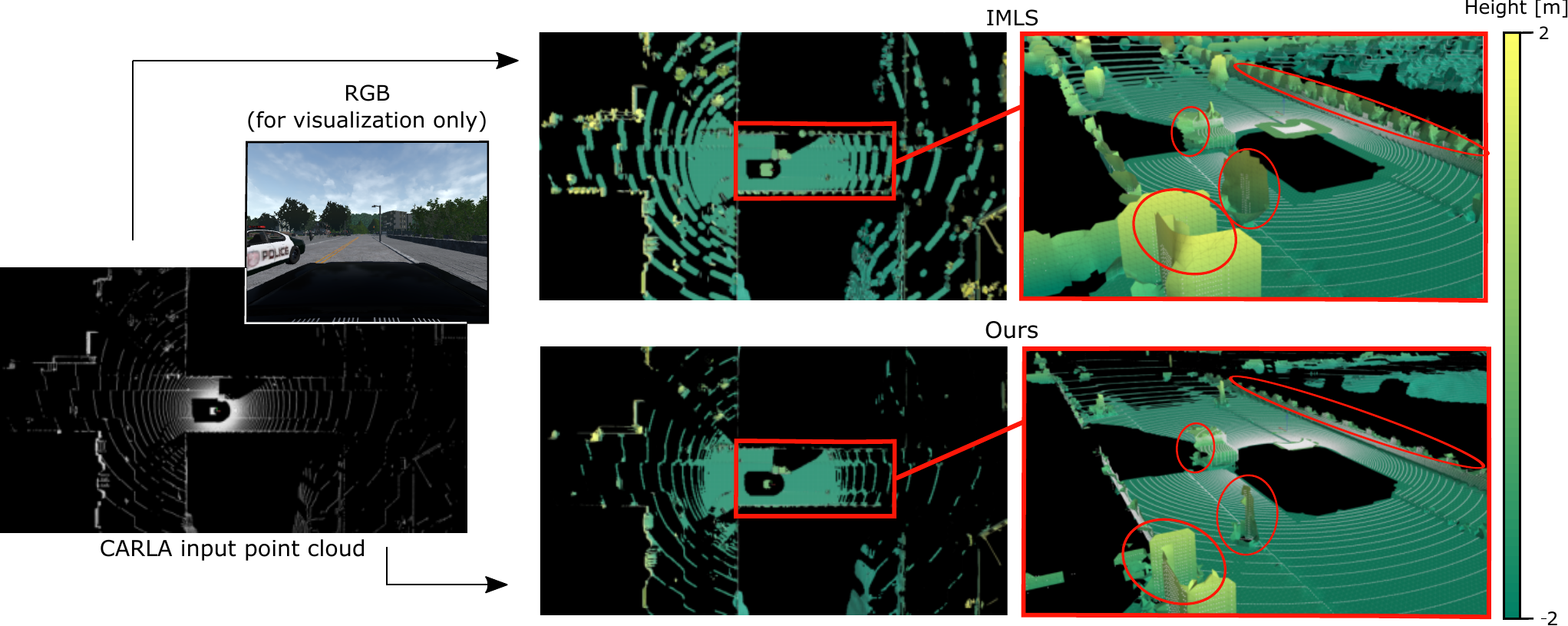}
	\caption{Qualitative comparison on CARLA simulator. Notice that even though IMLS outputs a denser reconstruction, it also extend all surfaces at its border and creates a higher number of artifacts as it can be seen by the red circles at rightmost images. Our method is able to keep the structure of the surface and generates fewer artifacts, performing a more accurate reconstruction, while keeping a good density in areas near to the vehicle.}
	\label{fig:comparison_carla}
\end{figure*}

\begin{figure}
	\centering
	\subfloat[Average error]{\includegraphics[width=0.48\columnwidth]{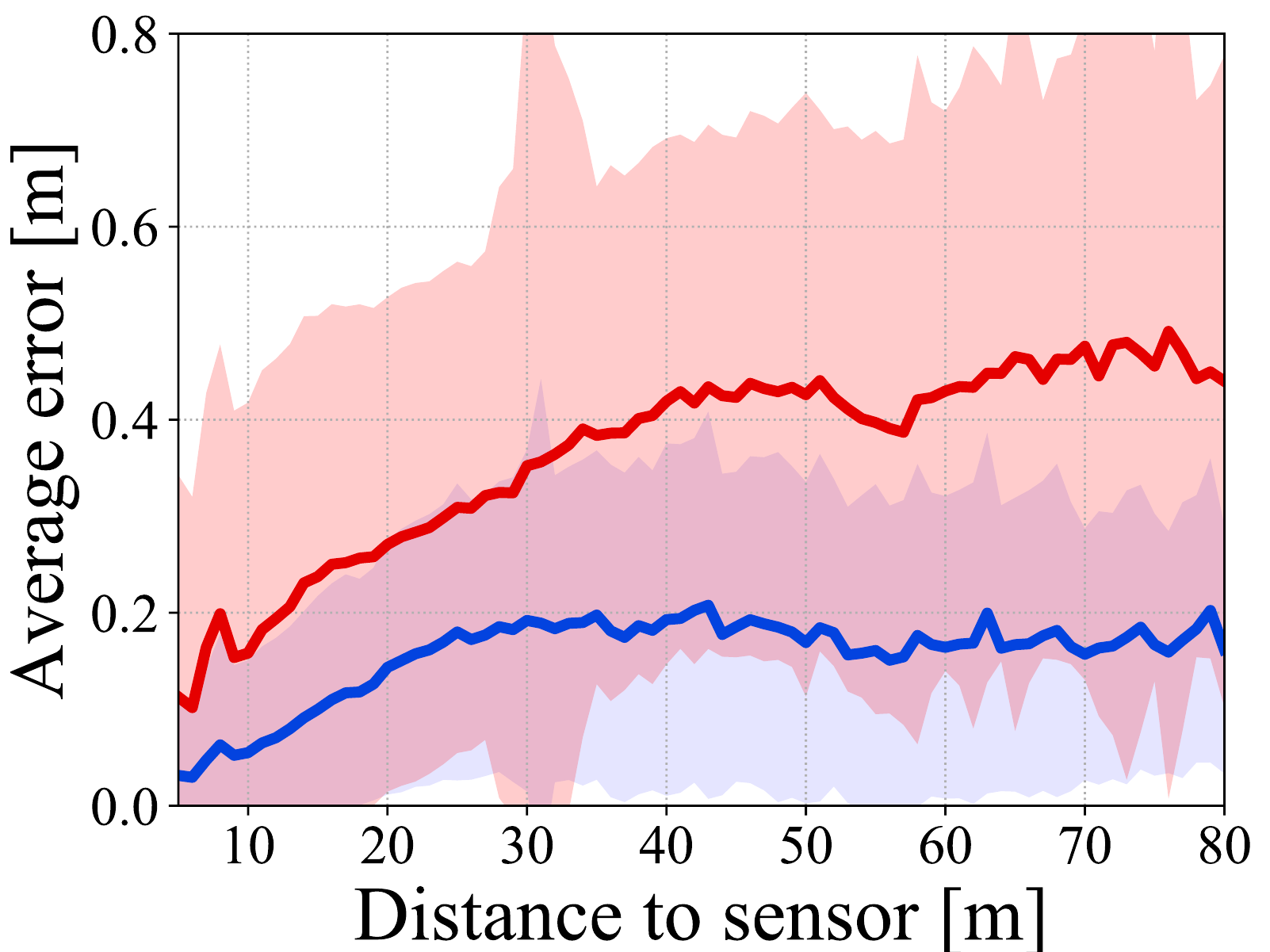}\label{fig:average_error_plot}}\hspace{0.01\columnwidth}\subfloat[Delta error]{\includegraphics[width=0.49\columnwidth]{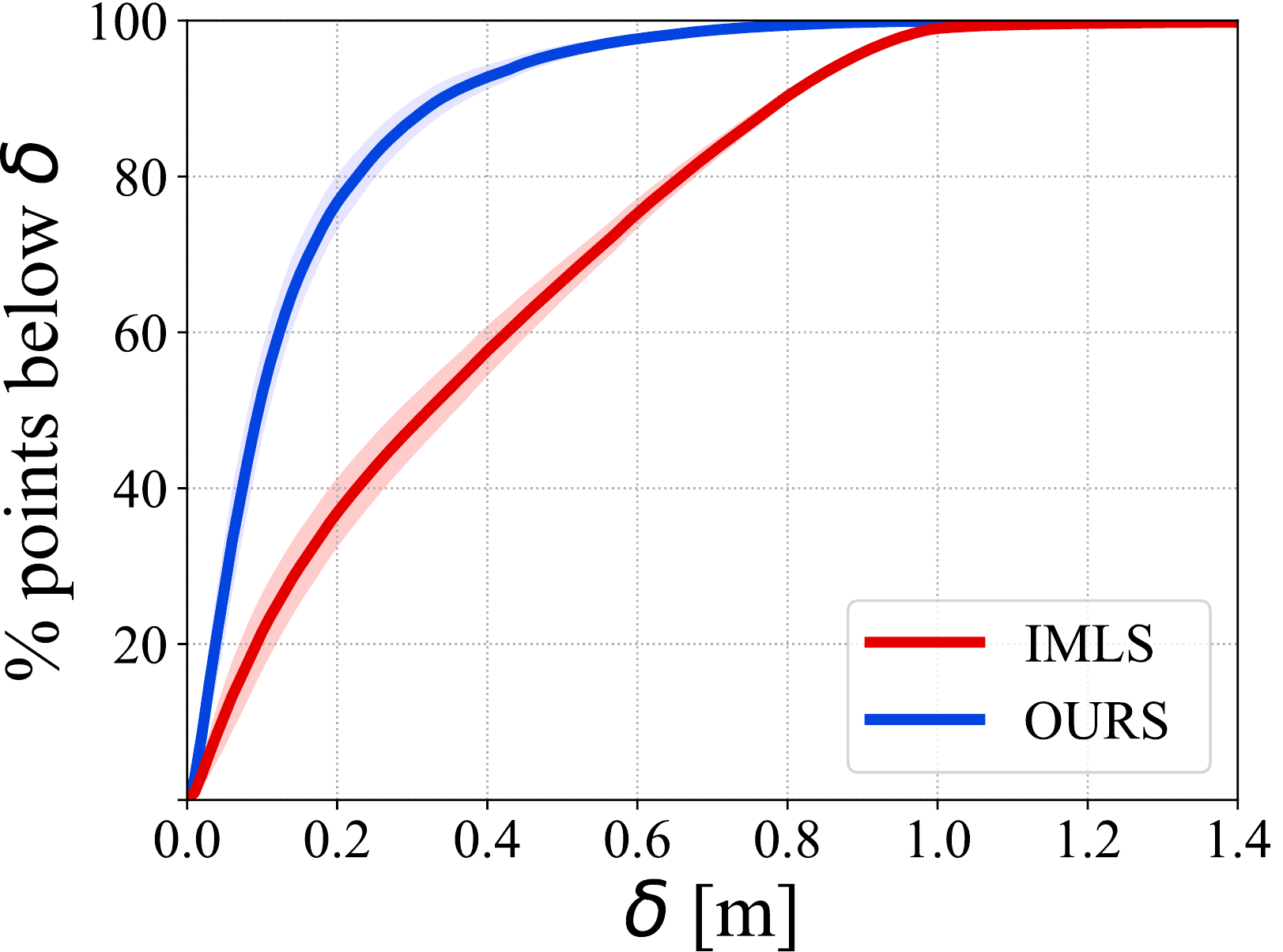}\label{fig:delta_error_plot}}
	\caption{Performance on the CARLA dataset from averaging of 100 frames evaluation. While the average error grows with the distance, at 30m distance our surface reconstruction error is $\approx0.2m$ whereas IMLS~\cite{Shen2004} is $\approx0.35m$.}
	\label{fig:error_plots}
\end{figure}

\subsection{Ablation study}\label{subsec:ablation_study}

We assess the average performance on 100 frames from CARLA and subsequently report quantitative performance in Table~\ref{tab:ablation_results} and qualitative results in Fig. \ref{fig:ablation_comparison}. In the former, our pipeline is compared to IMLS~\cite{Shen2004}. 

To evaluate the importance of our contributions, we compare the benefits of our Adaptive Neighborhood (\textbf{AN}, Sec.~\ref{subsec:adaptive_neighborhood}) with or without the Gaussian Confidence (\textbf{GC}, Eq.~\ref{eq:pdf_value}) and also with a naive Constant Neighborhood (\textbf{CN}) approach. More precisely, the neighborhood $\mathcal{H}^{k'}$ considered for the plane estimation (cf. Sec \ref{subsec:planar_estimation}) is different:
\\
\textbf{AN+GC:} $\mathcal{H}^{k'}$ is the minimum neighborhood among $\left\lbrace \mathcal{H}^{1},\cdots,\mathcal{H}^{k_{max}} \right\rbrace$ that satisfies Eq. \ref{eq:minimum_number_points} and \ref{eq:pdf_value}.
\\
\textbf{AN:} $\mathcal{H}^{k'}$ is the minimum neighborhood among $\left\lbrace \mathcal{H}^{1},\cdots,\mathcal{H}^{k_{max}} \right\rbrace$ that satisfies Eq. \ref{eq:minimum_number_points}.
\\
\textbf{CN+GC:}  $\mathcal{H}^{k'}$ with $k'=k$, is considered only if Eq. \ref{eq:minimum_number_points} and \ref{eq:pdf_value} are satisfied.
\\
\textbf{CN:} $\mathcal{H}^{k'}$ with $k'=k$, is considered only if Eq. \ref{eq:minimum_number_points} is satisfied.

From Table~\ref{tab:ablation_results}, the accuracy on the reconstruction directly affects the P$\rightarrow$GT distances while the GT$\rightarrow$P distances are mostly influenced by the reconstruction density. 

As expected, the benefit of our adaptive neighborhood strategy is noticeable when comparing \textit{AN+GC} and \textit{CN+GC}. Not only \textit{AN+GC} exhibits higher accuracy (lower P$\rightarrow$GT) but it also increases the reconstruction density (lower GT$\rightarrow$P). When using our Gaussian Confidence (GC), there is a slight density loss (lower GT$\rightarrow$P) but a significant accuracy increase.
Qualitatively from Fig. \ref{fig:ablation_comparison}, our adaptive neighborhood helps to maintain the details of the surface by not over-smoothing the data, while the Gaussian Confidence strategy avoids to extend the surface, keeping its structure and generating fewer artifacts.

Overall, we observe that our complete pipeline (AN+GC with $k_{\text{max}=5}$) keeps a good trade-off between accuracy and density, which is shown by the best result obtained $AE_{\text{sym}}$ ($0.14m$ vs. $0.23m$ for IMLS). 
With larger neighborhood (bigger $k_{\text{max}}$), the density of the reconstruction increases but at the expense of a lower accuracy which is intuitive as there is a need to extrapolate more the data.
Since IMLS performs a denser reconstruction, lower GT$\rightarrow$P distances are obtained, which explains the best results on the symmetric Hausdorff Distance ($10.94m$ vs. $6.43m$ for IMLS). While our method is less dense, our predicted mesh is significantly more precise ($1.39m$ vs. $4.54m$ for IMLS).

\begin{figure*}
	\centering
	\includegraphics[width=0.83\textwidth]{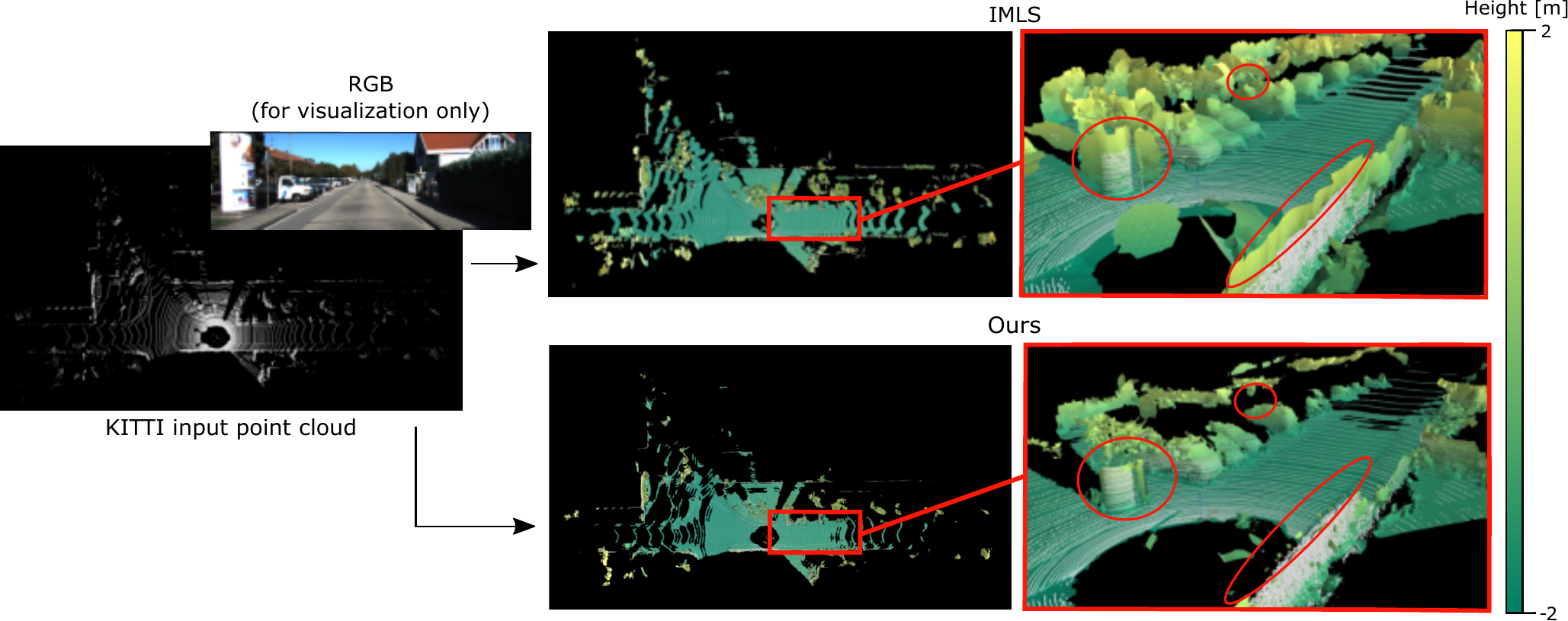}
	\caption{Visual comparison on KITTI dataset. The results on both methods show the same behavior than seen in synthetic data. IMLS performs a denser reconstruction at expenses of accuracy. Our method keeps a good trade-off between density and accuracy on the reconstruction.}
	\label{fig:comparison_kitti}
\end{figure*}

\subsection{Synthetic data}\label{subsec:syinthetic_data}

To further evaluate the performance, we report the average error as a function of the sensor distance in Fig.\ref{fig:average_error_plot}, and the cumulative delta error that indicates the percentage of vertices of the output meshes having an error lower than a given value in Fig. \ref{fig:delta_error_plot}. 

For both metrics, the accuracy of our reconstruction is significantly higher than the one obtained by IMLS, as the average error of our reconstruction is often 50\% lower (Fig.~\ref{fig:average_error_plot}). The percentage of vertices below a given error exhibits the same behavior (Fig. \ref{fig:delta_error_plot}), with a significant advantage for our method. Furthermore, almost $80\%$ of the vertices of our mesh have an error lower than $0.2m$, while $40\%$ of vertices lie below the same threshold with IMLS.

Finally, a qualitative comparison on CARLA is shown in Fig. \ref{fig:comparison_carla}. Again, one can see that IMLS outputs a more dense reconstruction of the scene but also that the method tends to extend all surfaces, generating artifacts and inaccuracies in the reconstruction. This can be observed in the circled areas where lampposts, traffic signs and walls are abnormally enlarged by IMLS. 
On the other hand, our method is able to keep these details on the reconstruction and maintains a high density in areas close to the vehicle, which confirms the quantitative results obtained in Fig.~\ref{fig:error_plots}.

\subsection{Real data}\label{subsec:real_data}

Qualitative results on KITTI residential sequence 0018 are shown in Fig. \ref{fig:comparison_kitti}. 
As for the results presented on the synthetic data, IMLS outputs a denser reconstruction but extends all surfaces out of the borders and generates artifacts. This can be observed at the circled areas on the rightmost images. Conversely, our method outputs a more accurate reconstruction, keeping the structure of the scanned surface and completing the surface at no data points areas close to the sensor. Even though our method outputs a less dense reconstruction, the density is still adequate for further robotics tasks such as trajectory planning or obstacle avoidance. Further quantitative results might be obtained by randomly sub-sampling the input data and calculating the reconstruction error over the non sub-sampled mesh.

\section{Conclusion}\label{sec:conlcusion}

In this paper we presented a surface reconstruction method based on a light voxel representation that benefits of statistical information.
Our pipeline uses an adaptive neighborhood strategy coupled with a Gaussian confidence estimation, to best estimate local surfaces and the implicit surface estimation (TSDFs).

The performance of our method was demonstrated on both synthetic (CARLA) and real data (KITTI), significantly outperforming the classical IMLS though sparser reconstruction. While the approach is computationally too expensive for online applications, it can be useful for terrain mapping or off road terrain analysis.

For mobile robots, an accurate representation of the environment is a prerequisite. Our method contributes to improve the reconstruction accuracy, which can be useful for mapping and localization applications.

While our results show that local planar surfaces are sufficient for accurate reconstruction, we intend to extend this research to a more complex surface definition, for example with polynomials.

\bibliographystyle{IEEEtran}
\bibliography{References}

\end{document}